\newif\ifusenix
\newif\ifacm
\newif\ifmcom
\newif\ifieee

\usenixfalse
\acmfalse
\mcomfalse
\ieeetrue

\ifieee
    \documentclass[letterpaper, 10 pt, conference]{ieeeconf}  %
    
    \IEEEoverridecommandlockouts                              %
\fi

\ifusenix
        \documentclass[letterpaper,twocolumn,10pt]{article}
        \usepackage{usenix2019_v3}

        \usepackage{times}
\fi

\ifacm
  \documentclass[sigconf,10pt]{acmart}
  \renewcommand\footnotetextcopyrightpermission[1]{} %

\fi

\ifmcom
  \documentclass{sig-alternate-10pt}
\fi

\usepackage{xcolor}
\usepackage{placeins}
\usepackage{amsfonts}
\usepackage{balance}
\PassOptionsToPackage{hyphens}{url}\usepackage{hyperref}
\usepackage{color}
\usepackage{graphics}
\usepackage{graphicx}
\usepackage{url}
\usepackage{listings}
\usepackage{multicol}
\usepackage{multirow}
\usepackage[scaled]{helvet}
\usepackage{rotating}
\usepackage{xspace}
\usepackage[ruled,vlined]{algorithm2e}
\usepackage{comment}
\usepackage{amsmath}
\usepackage{mathrsfs}
\usepackage{cancel}
\usepackage{textgreek}
\usepackage{authblk}
\usepackage[font=small,labelfont=bf]{caption}
\usepackage[skip=0pt]{subcaption}
\usepackage{enumitem}
\setlist{nosep,noitemsep,topsep=0pt,parsep=0pt,partopsep=0pt,leftmargin=*,parsep=0pt}

\usepackage{microtype}
\usepackage{multirow}

\usepackage[capitalise]{cleveref}
\crefname{section}{Sec.}{Sec.} %
\crefname{algocf}{Alg.}{Algs.}

\newcommand{\todo}[1]{\ClassWarning{NOT READY TO SUBMIT}{There is something left todo} \textcolor{blue}{}}

\newcommand{\name}{WiROS\xspace}
\renewcommand{\iota}{\textsl{j}\xspace}

\usepackage{etoolbox}
\apptocmd\normalsize{%
 \abovedisplayskip=2pt plus 2pt minus 2pt
 \abovedisplayshortskip=2pt plus 2pt
 \belowdisplayskip=2pt plus 2pt minus 2pt
 \belowdisplayshortskip=2pt plus 2pt minus 2pt
}{}{}

\usepackage[skip=0pt]{subcaption}
\usepackage[explicit,noindentafter]{titlesec}
\titlespacing\section{3pt}{4pt plus 2pt minus 2pt}{1pt plus 2pt minus 2pt}
\titlespacing\subsection{3pt}{4pt plus 2pt minus 2pt}{1pt plus 2pt minus 2pt}
\titlespacing\subsubsection{3pt}{4pt plus 2pt minus 2pt}{1pt plus 2pt minus 2pt}

\title{
\name: WiFi sensing toolbox for robotics 
}

\author{Submission: \#xxx, 12 pages}
\author{William Hunter$^{*1}$, Aditya Arun$^{*2}$ and Dinesh Bharadia$^{3}$ \\ *equal contribution%
\thanks{$^{1}$William Hunter is with Electrical and Computer Engineering Dept.,
        University of California, San Diego, CA 92092, USA
        {\tt\small wshunter@eng.ucsd.edu}}
\thanks{$^{2}$Aditya Arun is with Electrical and Computer Engineering Dept.,
        University of California, San Diego, CA 92092, USA
        {\tt\small aarun@ucsd.edu}}%
\thanks{$^{2}$Dinesh Bharadia is faculty with Electrical and Computer Engineering Dept., 
        University of California, San Diego, CA 92092, USA
        {\tt\small dineshb@eng.ucsd.edu}}%
}
        
\ifacm
    \begin{abstract}
Many recent works have explored using WiFi-based sensing to improve SLAM~\cite{arun2022p2slam}, robot manipulation~\cite{boroushaki2022fusebot} or exploration~\cite{clark2022propem}. Moreover, widespread availability makes WiFi the most advantageous RF signal to leverage. But WiFi sensors lack an accurate, tractable, and versatile toolbox, which hinders their widespread adoption with robot's sensor stacks. 

We develop \name to address this immediate need, furnishing many WiFi-related measurements as easy-to-consume ROS topics. Specifically, \name is a plug-and-play WiFi sensing toolbox providing access to coarse-grained WiFi signal strength (RSSI), fine-grained WiFi channel state information (CSI), and other MAC-layer information (device address, packet id's or frequency-channel information). Additionally, \name open-sources state-of-art algorithms to calibrate and process WiFi measurements to furnish accurate bearing information for received WiFi signals. The open-sourced repository is: \href{https://github.com/ucsdwcsng/WiROS}{\name}:https://github.com/ucsdwcsng/WiROS. 
   
\end{abstract}

\fi

\begin{document}
\maketitle

\ifieee
    \begin{abstract}
Many recent works have explored using WiFi-based sensing to improve SLAM~\cite{arun2022p2slam}, robot manipulation~\cite{boroushaki2022fusebot} or exploration~\cite{clark2022propem}. Moreover, widespread availability makes WiFi the most advantageous RF signal to leverage. But WiFi sensors lack an accurate, tractable, and versatile toolbox, which hinders their widespread adoption with robot's sensor stacks. 

We develop \name to address this immediate need, furnishing many WiFi-related measurements as easy-to-consume ROS topics. Specifically, \name is a plug-and-play WiFi sensing toolbox providing access to coarse-grained WiFi signal strength (RSSI), fine-grained WiFi channel state information (CSI), and other MAC-layer information (device address, packet id's or frequency-channel information). Additionally, \name open-sources state-of-art algorithms to calibrate and process WiFi measurements to furnish accurate bearing information for received WiFi signals. The open-sourced repository is: \href{https://github.com/ucsdwcsng/WiROS}{\name}:https://github.com/ucsdwcsng/WiROS. 
   
\end{abstract}

\fi

\section{Introduction}

Incorporating wireless sensing into a robotics sensor stack allows robots to better handle the failure cases of visual sensors like cameras or Lidars. Many recent works~\cite{arun2022p2slam, boroushaki2022fusebot, zouGAN2022} have shown that integrating wireless signals helps to overcome common failure cases like visual occlusion, dynamic lighting, or perceptual aliasing~\cite{lajoie2019modeling}. Among the various wireless-sensing modalities explored, leveraging WiFi sensing is particularly attractive. It is widely deployed in many indoor environments, most robots already have WiFi chipsets for communication purposes, and it provides an extended sensing range (over 10 m). With this in mind, many recent papers~\cite{arun2022p2slam, zouGAN2022, hashemifar2019augmenting} explore using WiFi signals. 

However, no easy solutions exist for integrating them into an existing robot's sensor stack, precluding their wider use. Popular cameras~\cite{realsense} and Lidars~\cite{hoku} enjoy wide support within the Robot Operating System (ROS) framework with clear documentation. This ROS support allows for quick integration and easy operation of these sensors. The same, however, cannot be said for WiFi sensors, inhibiting their adoption for robotics applications. In this work, we present \name to fill in this gap. \name develops a set of ROS nodes that furnishes raw WiFi measurements, processed WiFi features, and visualizations as ROS topics and provides an easy-to-perform wireless calibration framework. 

With the broad goal of providing accessible WiFi sensing information within robot sensor stacks, we follow three design principles. A WiFi sensor wrapper should be 
\begin{enumerate}
    \item \textbf{Accurate}: provide WiFi sensing measurements accurately,
    \item \textbf{Tractable}: quick to bring up and easy to calibrate,   
    \item \textbf{Versatile}: widely usable in WiFi-equipped spaces.
\end{enumerate}

\begin{figure}[t]
    \centering
    \includegraphics[width=0.9\linewidth]{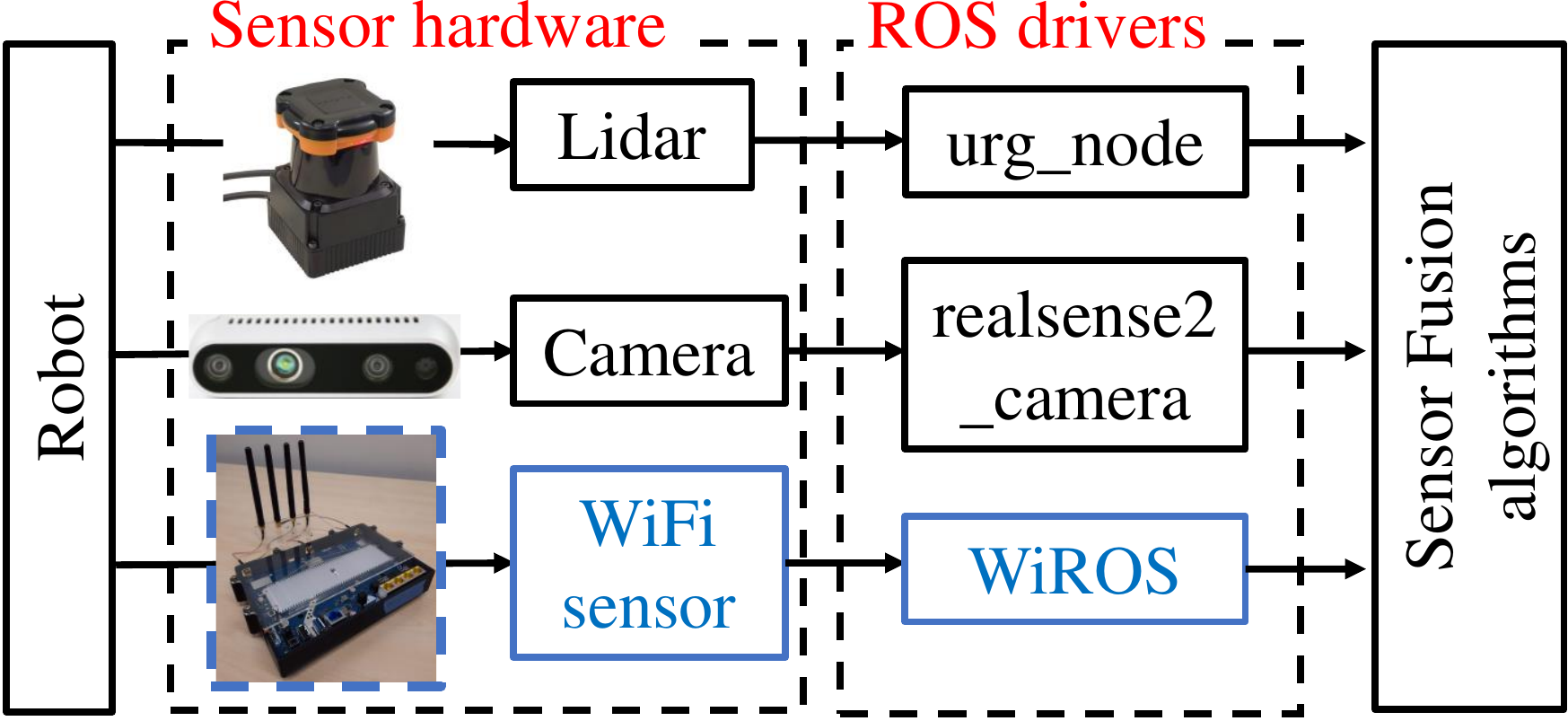}
    \caption{The sensor hardware, like cameras and Lidars, have widely supported ROS nodes. This work releases \name, a similar ROS node compatible with WiFi sensors.}
    \label{fig:intro_fig}
\end{figure}

\begin{figure}[t]
    \centering
    \includegraphics[width=0.69\linewidth]{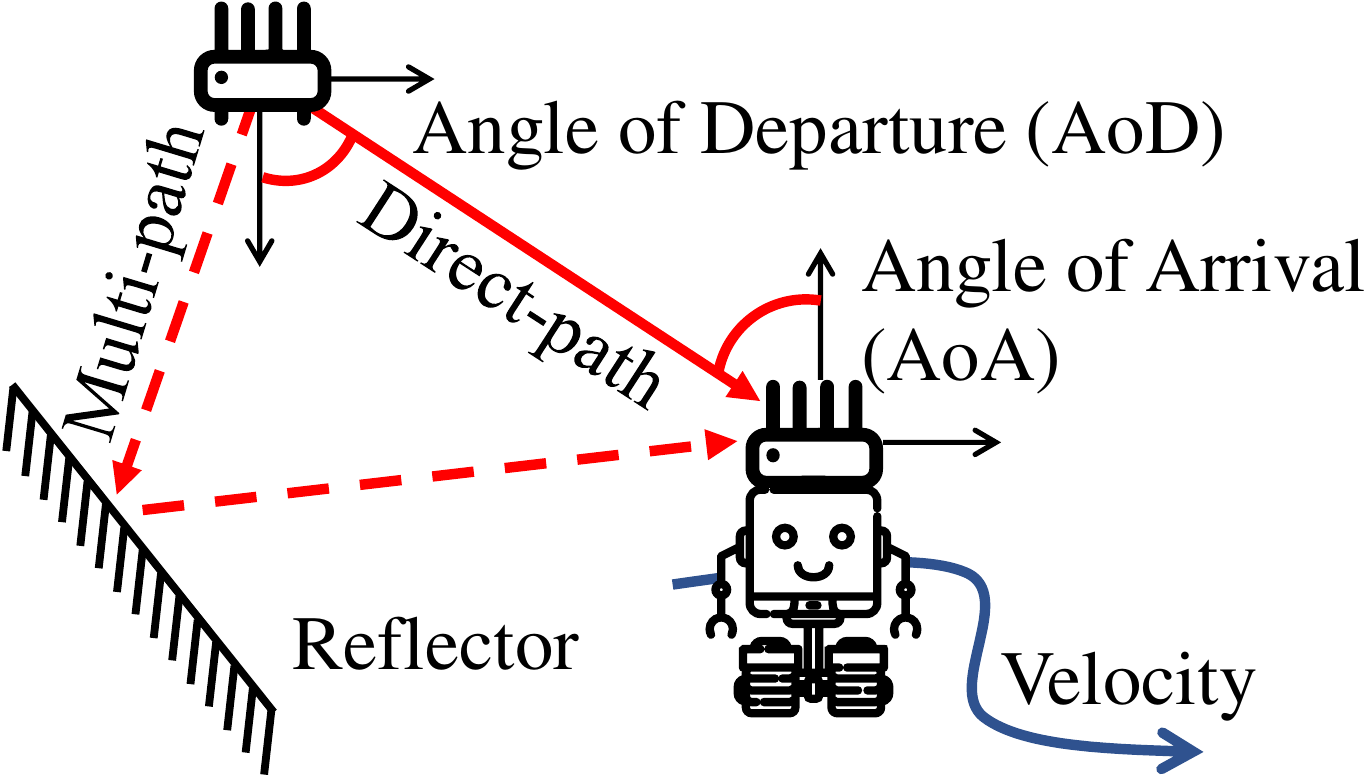}
    \caption{\textbf{Signal parameters measurable by CSI} The direct path's (solid line) and multi-path's (dotted line) angles of arrival and departure can be measured in the local coordinates of the APs. The robot's velocity may also be measured. }
    \label{fig:bg-info}
\end{figure}

\noindent\textbf{\underline{Prior Works}:} However, many existing systems that furnish WiFi measurements at the robot application layer fail to meet one or a few of these requirements, making them unsuitable for robot integration.
\begin{itemize}
    \item \textbf{WiFi measurement tools}: There are two widely used open-source WiFi measurement toolkits~\cite{halperin2011tool, atheroscsi} which support the IEEE 802.11n protocol. However, this protocol is outdated and consequently not as widely supported in current WiFi deployments. Alternatively, newer toolkits that leverage the IEEE 802.11ac~\cite{gringoli2019free} and IEEE 802.11ax~\cite{gringoli2022ax, jiang2021eliminating} exist. Unfortunately, these systems do not expose WiFi measurements in real-time, instead storing it on the device in specialized files. This requires additional post-processing and time-synchronization, precluding real-time robot operation. Consequently, these toolkits are neither versatile nor tractable. 
    \item \textbf{ROS-supported Tools}: To circumvent these challenges, a few tools integrate WiFi measurements into ROS frameworks. \cite{wifitool} provides only the WiFi signal strength (RSSI) measurements via a ROS topic, which can be used as a proxy between a transmitter and receiver. However, RSSI is a very coarse-grained, environment-sensitive measurement, whereas richer WiFi measurements exist that can expand the use of WiFi sensors. These richer measurements (channel state information, CSI) are exposed in the tools above and help determine the arrival and departure angles of a WiFi signal, the velocity of the WiFi sensor, or even fine-grained information about the environment~\cite{ma2019wifi}. Figure~\ref{fig:bg-info} details these physical measurements. A recent work~\cite{jadhav2022toolbox} looked at providing ROS support for the 802.11n chipsets~\cite{halperin2011tool, atheroscsi} mentioned previously. However, this system requires active collaboration with the WiFi infrastructure, requiring deployed WiFi APs to perform `round-robin' packet exchanges. This infrastructure dependency precludes ubiquitous deployment of WiFi sensing in indoor environments and presents logistic, security, and networking challenges. For instance, a public deployment of their system would require firmware upgrades on deployed WiFi access points, authentication of third-party robots with a secure network, and introduce additional burdens on a congested network. Additionally, as of this writing\footnote{Referring to \href{https://github.com/Harvard-REACT/WSR-Toolbox/tree/7e795086ccd4c11935e2685080f62848e18801c2}{Github commit 7e79508}}, it does not provide ready integration to ROS. 
\end{itemize}

\name instead leverages widely used 802.11ac WiFi protocol, which provides $2\times$ the sensing bandwidth (improving \textit{accuracy}). We build on prior work~\cite{gringoli2019free} a WiFi sensing framework making the following  contributions and overcoming the limitation of the prior works:

\noindent \textbf{1. Scalable framework for WiFi sensing:} We provide a ROS node (`CSI-Node') as a simple abstraction over the hardware used to collect CSI to allow for simple integration and leverage ROS-based time synchronization. \name subsequently extracts various physical parameters (angles of arrival and departure of the signal, angular positions of reflections in the environment) to aid robot operation. For example, \name can measure the bearings of other robots or arbitrary transmitters (WiFi APs, laptops, phones, or IoT devices) from a single received WiFi packet, without associating or authenticating with the existing network. This prevents network congestion, does not require collaboration with network infrastructure, and allows for ubiquitous deployment of our system in multiple indoor spaces. Clearly, \name enables \textit{versatile} WiFi sensing, and additional details can be found in Section~\ref{sec:extend}. 

\noindent \textbf{2. Easy calibration and setup:} Commercial APs come with unique hardware biases, which can skew the signal measurements~\cite{roarray}. These biases need to be measured apriori, and measurements must be calibrated to estimate various physical parameters, including the bearing of the transmitted signal. Past works~\cite{ubilocate, xiong2010secureangle, jadhav2022toolbox} have required disassembly of the device and manual measurement of these biases, severely reducing the tractability of a sensor platform. Instead, we provide a solution for automatically calibrating the phase offsets on-robot by extending the work in \cite{roarray}. This allows for hassle-free and real-time wireless calibration, with Section~\ref{sec:calib} providing further details.

\noindent \textbf{3. Algorithms for sensing:} Finally, we provide a ROS node (`Feature-extraction Node') to estimate the angles of arrival and departure of a WiFi signal using state-of-the-art techniques~\cite{spotfi, schmidtmusic}. Additionally, we provide intuitive visualizations of the received WiFi signal to aid debugging. Providing this node serves two key purposes. First, it allows out-of-the-box use of WiFi measurements for SLAM and other navigation purposes. Second, it provides a blueprint for using WiFi CSI to perform wider tasks like Doppler estimation or time-of-flight measurements~\cite{xie2019md}. These concepts are further elaborated in Section~\ref{sec:bearing}.

In the next section, we will provide a high-level overview of \name's usage (Section~\ref{sec:usage}) and specific details about the above three contributions. 

\section{Design and Usage} \label{sec:design}

\begin{figure*}[t]
    \centering
    \includegraphics[width=0.93\linewidth]{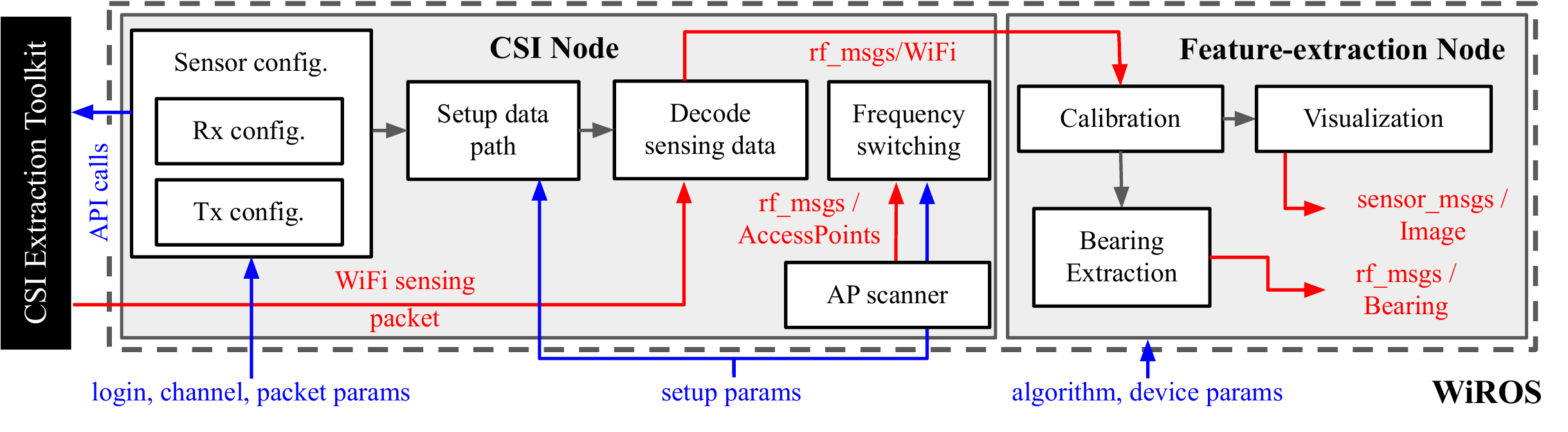}
    \caption{\textbf{\name's Block Diagram}: An inside look at the \name's box from Fig~\ref{fig:intro_fig}. Showcases the two main blocks - CSI Node and Feature-extraction Node to extract raw WiFi measurements and to calibrate and process these measurements. The blue text indicates the control plane parameters, whereas the red text indicates the exposed measurements. \name extends the functionality of the underlying black box `CSI Extraction Toolkit.'} 
    \label{fig:pipeline}
\end{figure*}

\name's primary motivation is to provide an accessible WiFi sensor within robot sensor stacks to leverage key advantages WiFi signals provide. In this vein, we will first provide a quick rundown on the usage of \name, the WiFi measurements exposed via ROS topics and visualizations of these measurements (Section~\ref{sec:usage}). This will act as an overview for the extensive documentation provided along with the code repositories\footnote{\label{fn:links} \name's Index page: \href{https://github.com/ucsdwcsng/WiROS}{https://github.com/ucsdwcsng/WiROS} \\ Sub-links: CSI Node: \href{https://github.com/ucsdwcsng/wiros_csi_node}{https://github.com/ucsdwcsng/wiros\_csi\_node} \\ Feature-extraction Node: \href{https://github.com/ucsdwcsng/wiros_processing_node}{https://github.com/ucsdwcsng/wiros\_processing\_node} \\ Custom RF messages: \href{https://github.com/ucsdwcsng/rf_msgs}{https://github.com/ucsdwcsng/rf\_msgs} \\}

\name's secondary motivation is to provide a replicable framework that can be easily integrated with different WiFi radio hardware and CSI extraction toolkits. In this vein, we will provide details of

\noindent\textbf{(a)} extending Nexmon-CSI~\cite{gringoli2019free} as an example in developing \name (Section~\ref{sec:extend}), 

\noindent\textbf{(b)} the wireless calibration techniques incorporated within \name (Section~\ref{sec:calib}), and 

\noindent\textbf{(c)} the open-sourced implementation of various state-of-art bearing estimation algorithms for readily using WiFi measurements (Section~\ref{sec:bearing}).

\subsection{Using \name} \label{sec:usage}

First, we demonstrate usage of WiROS with Asus RT-AC86U~\cite{asus}, however, other chipsets can also be readily used, and we provide further details on how one can integrate other chipsets into \name's framework in Section~\ref{sec:other}. To setup the hardware, we need to follow three simple steps
\begin{enumerate}
    \item Compile the open-source ROS packages using \texttt{catkin}.$^{\ref{fn:links}}$
    \item Setup the access point by flashing the provided firmware.
    \item Configure provided ROS \texttt{params} and stream the WiFi measurements as ROS \texttt{topics}.
\end{enumerate}
To elaborate on these steps, we will set up this Asus off-the-shelf access point as both a WiFi receiver or a transmitter using \name's CSI Node$^{\ref{fn:links}}$. As shown in Figure~\ref{fig:pipeline}, the CSI Node takes as input ROS \texttt{params}, in blue, which will be discussed briefly below. 

\noindent\textbf{Setting up a receiver:} The specific WiFi frequency-channel and bandwidth should be first configured via  `channel' params. A filter for specific MAC addresses can also be included via the `packet' params. If an all-pass filter is used, \name will monitor and provide WiFi measurements for received signals for all WiFi devices transmitting on the specified WiFi channel and bandwidth. With these configurations set up, the receiver can be brought up with \texttt{rosrun wiros\_csi\_node csi\_node}. Alternatively, simple \textit{launch} files are provided to further improve \textit{tractability}. 

However, as mentioned, receiving WiFi signals from a specific channel and bandwidth can reduce usability in a wild environment. Most enterprise WiFi networks deploy multiple access points (AP) that serve users across a building. Neighboring APs in these networks are often deployed on different frequency channels to reduce interference. Hence, \name will miss out on CSI measurements from APs configured on other channels. To ensure \textit{versatility} of our system, we implement a ROS \texttt{service} which allows a user to switch channels on the WiFi sensors. Additionally, \name automates this channel switching to monitor the channel of the nearest AP by periodically scanning all the WiFi channels. This automated behavior can be enabled via the `setup' params.

\noindent\textbf{Setting up a transmitter:} Additionally, researchers may often require a steady source of WiFi transmissions, with configurations of the WiFi frequency-channel, bandwidth or MAC address. For example, later in Section~\ref{sec:calib}, we will need to configure a WiFi AP to transmit packets continuously to compute the hardware calibration parameters. The `channel' and `setup' params can be used to configure the transmitter. With these parameters configured, the AP can be brought up with the same \texttt{rosrun} command as before. Next, we will discuss an overview of the exposed measurements available as ROS \textit{topics} via an Ethernet connection. 

\noindent\textbf{Exposed measurements:} Once a receiver has been set up, it will capture WiFi measurements from ambient WiFi signals in the configured frequency-channel and bandwidth. These measurements will be available via a ROS \textit{topic} through a custom WiFi-sensing ROS \textit{message}, \texttt{rf\_msgs/Wifi} as shown in Figure~\ref{fig:pipeline}. Amongst the various exposed information, the signal strength (RSSI) and the channel state information (CSI) are the two commonly used measurements. RSSI, measured in dB, provides a rough estimate of the distance of the transmitter and is widely used as a proxy to measure communication throughput~\cite{azini2016transparent}, as a feature for improving SLAM performance~\cite{hashemifar2019augmenting} or to map the transmitters~\cite{sharma2022lumos} coarsely. Alternatively, CSI provides more fine-grained information. It is a matrix of complex numbers indicating the received signal magnitude and phase across a set of transmitted frequencies and receiver antennas. The number of transmitted frequencies is controlled by the bandwidth ($20$, $40$ or $80$ Mhz), with the $80$ MHz bandwidth only supported in the 802.11ac protocol. By default, for the Asus hardware, there are $4$ receive antennas compared to $3$ receive antennas commonly present in alternative systems~\cite{atheroscsi, halperin2011tool}. The larger bandwidth and number of antennas provide more \textit{accurate} sensing capabilities. Additionally, if a transmitter has multiple antennas configured, an additional dimension for the number of transmit antennas are present. However, these CSI measurements are challenging to visualize. To improve the \textit{tractability} of WiFi sensors, we provide further details on the real-time visualizations included in \name next.  

\begin{figure}
    \centering
    \centering
    \includegraphics[width=0.95\linewidth]{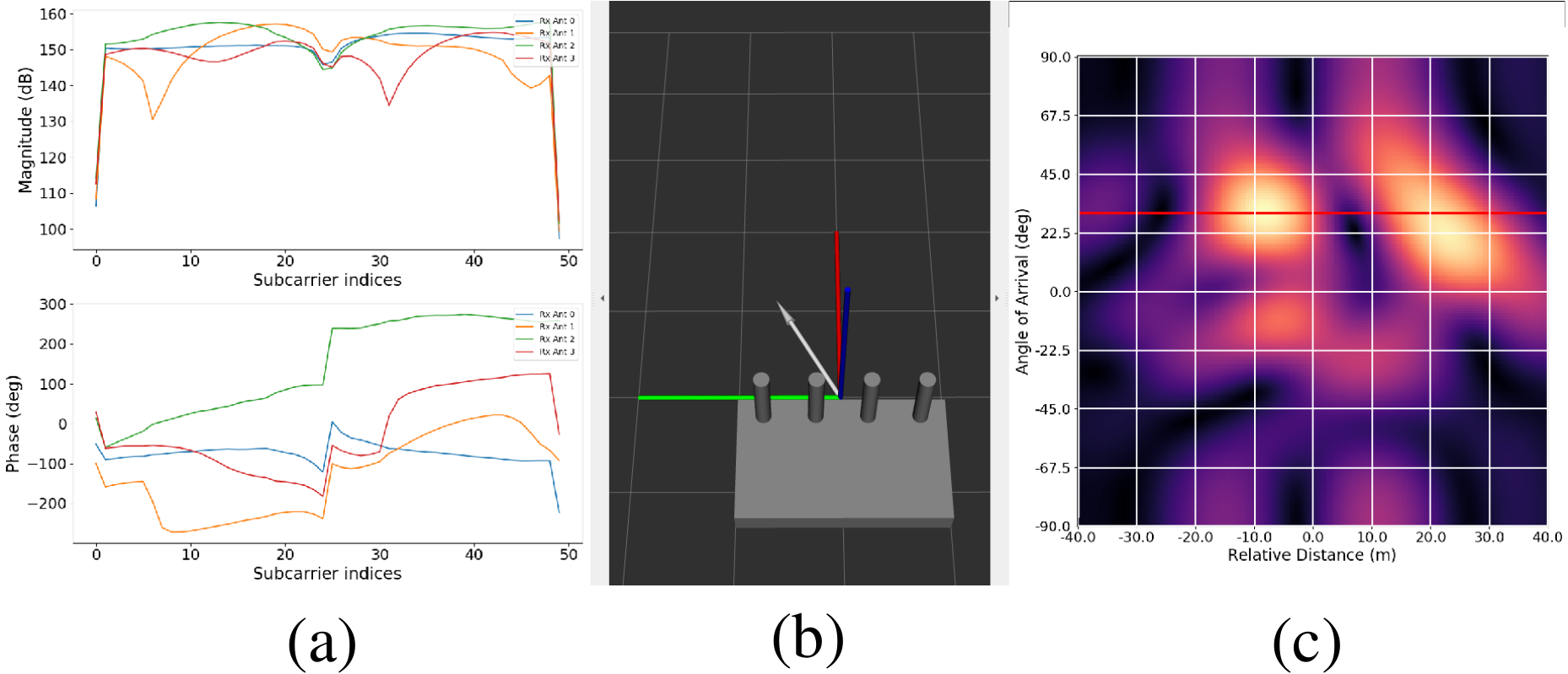}
    \caption{\textbf{(a)} \textbf{Magnitude-phase profile}, with the top plot showcasing the magnitude and bottom plot the phases across 4 receivers and 234 subcarrier frequencies. \textbf{(b)} A simplified visualization of the predicted bearing of the signal (white arrow) in the local coordinates of the access point. 
    \textbf{(c)} \textbf{Bearing-range likelihood profile}, with the red-line indicating the strongest received path, and hence the bearing of the signal. These images are visualized via the ROS \texttt{RViz} software in realtime}
    \label{fig:vis}
\end{figure}

\noindent\textbf{Visualization and processing measurements:} The CSI measurements available via ROS \texttt{topics} can be processed and visualized by deploying the Feature-extraction node$^{\ref{fn:links}}$. This node furnishes two important visualizations -- the magnitude-phase profile and the bearing-range likelihood profile. The magnitude-phase profile (see Figure~\ref{fig:vis}(a)) provides a visualization of the CSI measurements, providing the magnitude (y-axis of top image) and phase (y-axis of bottom image) of the received signal in decibels and degrees, respectively, across various subcarrier frequencies (in the x-axis) and across the four receive antennas (as different lines). The bearing-range profile (see Figure~\ref{fig:vis}(c)) indicates the arrival of a WiFi signal from the transmitter via a straight-line direct path and numerous reflected paths. The profile's peaks indicate the bearing of the transmitter (in the vertical axis) and the relative distances of the direct-path signal and its reflected-path echoes. Additionally, visualization of these profiles is provided in the supplementary video.\footnote{\label{fn:video-link} Youtube demo link: \href{https://youtu.be/zYAshWFn75k}{https://youtu.be/zYAshWFn75k}} Both these visualizations, exported as \texttt{sensor\_msgs/Image} as shown in Figure~\ref{fig:pipeline}, can be utilized to debug the wireless channel, check if all the receive antennas are performing equivalently or tune processing parameters like algorithms to leverage, time-window for averaging or RSSI thresholds. Additionally, a simple visualization of the received signal's bearing (white arrow) in the access point's local coordinates is shown in Figure~\ref{fig:vis}(b). 

Additionally, the Feature-extraction node open-sources state-of-art bearing estimation algorithms~\cite{spotfi, arun2022p2slam, schmidtmusic} to measure the angles of arrival and departure of the received signal. These measurements are processed in real-time and exposed as ROS \texttt{topics} via custom WiFi-bearing messages, \texttt{rf\_msgs/Bearing}. 

This brief description showcases the ease of use of \name's WiFi sensor for real-time data collection, processing, and debugging, improving state-of-the-art systems that require cumbersome data file post-processing. In addition to this description, extensive setup and usage documentation is provided with the code-repositories$^{\ref{fn:links}}$. The next section will provide further specifics about the implementation details and discuss the specific process block in Fig.~\ref{fig:pipeline}.

\subsection{CSI Node: Middleware to integrate with ROS} \label{sec:extend}

The `CSI node' is modularly written in C++, allowing it to interface with multiple CSI extraction toolkits (the black box in Figure~\ref{fig:pipeline}). It is currently tested with Nexmon-CSI's~\cite{gringoli2019free} Asus RT-AC86U~\cite{asus} platform as it is the most capable extant solution for COTS CSI extraction. The CSI Node is tested on ROS Kinetic, Melodic, and Noetic. It runs out-of-the-box on a Raspberry PI, or via a dedicated Ethernet connection to a central server.

\noindent\textbf{Data Path}: \name takes as input raw CSI data packets from the black box (shown in red arrows) via a UDP socket configured over an Ethernet connection by the `Setup Data-path' block. The `Decode Data' block decodes these raw WiFi data packets to expose the measurements in easy-to-consume \texttt{rf\_msgs/WiFi} format. 

\noindent\textbf{Control Path}: The CSI-node additionally provides different radio configurations via API calls. This abstracts out the specific hardware implementation for the end user. A dedicated `Sensor Config' submodule configures the underlying WiFi radio. It requires `login' params to access the WiFi AP, `channel' params to configure the WiFi frequency-channel and bandwidth, and `packet' params to configure the transmitter's beacon rate or filter WiFi data packets.  

Functioning in parallel, two sub-modules, if enabled via the `setup' params, are responsible for tracking and adjusting the frequency-channel used by the AP. The `AP scanner' periodically scans the different channels and determines the closest WiFi AP to re-configure the WiFi radio by exposing the information via the \texttt{rf\_msgs/AccessPoints} message. The `Channel switching' sub-module handles switching the channel with minimal sensing downtime (less than 500 milliseconds).

\subsection{Quick and easy calibration} \label{sec:calib}

Calibrating a sensor is a necessary first step and must be easy to perform. Generally, the calibration can vary for different frequency-channels and is unique for each hardware. This necessitates an easy-to-deploy and accurate calibration framework for the tractability of the WiFi sensor. This section will elaborate on the one-time wireless calibration system provided via a Python3 script in the `Feature-extraction' node. 

\noindent\textbf{Usage:} To calibrate our raw WiFi messages collected in the previous section, we must apply independent phase corrections across each antenna and frequency measurement. To compute this calibration, first, configure a receiver and transmitter to a specific WiFi frequency-channel as discussed in Section~\ref{sec:usage}. Next, collect raw CSI measurements by placing a WiFi sensor `receiver' on a robot and a WiFi sensor `transmitter' in a static predefined location in space. Instead of an ASUS WiFi sensor, a phone or laptop may be configured to transmit `ping' packets. In this setup, we are looking to calibrate the WiFi sensor receiver. Run the robot in any pattern in relatively free space, within a 5 m radius of the transmitter, and collect the robot odometry measurements ($\vec{r}_t$) as \texttt{nav\_msgs/Odometry} and the WiFi measurements ($W_t$) as \texttt{rf\_msgs/Wifi} from the CSI node. Note the location of the transmitter ($\vec{t}$) in the robot's generated map (often visible when using a LiDAR) and measure the relative antenna locations ($\vec{a}_i$) on the receiver. This data can be input into the calibration framework to generate the wireless calibration matrix, provided with the `Feature-extraction' Node$^{\ref{fn:links}}$.  

\noindent\textbf{Behind the scenes:} From our discussion so far, we need to find a phase calibration matrix $C$, 
$$ C = \exp(j \Phi) \in \mathbb{C}^{4 \times N_f},$$ 
to calibrate the phase measurements across the 4 antennas and $N_f$ frequencies, where $\Phi \in \mathbb{R}^{4 \times N_f}$. Given a raw CSI measurement from the CSI Node, $W_t \in \mathbb{C}^{4 \times N_f}$, the calibration is applied as 
$$ W^\mathrm{cal}_t = C \odot W_t, $$
where $\odot$ is the Hadamard (element-wise) product. 

Using the robot poses ($\vec{r}_t = (r_t^x, r_t^y, r_t^\theta) \in SE(2)$) and transmitter location ($\vec{t} \in \mathbb{R}^2$), we first compute the expected ground truth bearings ($\theta_t$). These can then be converted to expected WiFi CSI measurements ($\hat{W}_t \in \mathbb{C}^{4 \times N_f}$, implicitly assuming 4 receive antennas). 

\begin{align*}
    \theta_t &= \frac{\pi}{2} - \left(\arctan\left(\frac{r_t^y - t^y}{r_t^x - t^x}\right) - r_t^\theta\right) \\
    \hat{W}^{i, j}_t &= \exp\left(\frac{2\pi j}{\lambda} [\cos(\theta_t),\sin(\theta_t)] \vec{a}_i\right) \ \forall j \in [ 1, N_f],\forall i \in [1, 4]
\end{align*}
where, $\vec{a}_i$ is the relative location of the $i^\mathrm{th}$ antenna with respect to the first antenna; consequently, $\vec{a}_1 = \vec{0}$. $\lambda$ is the wavelength of the center frequency for the WiFi channel in consideration. Assuming a strong line-of-sight path signal is present in our measurements, we can expect the phases $\angle\hat{W}_t \approx \angle W^\mathrm{cal}_t$. Note that our assumption is reasonable given that the calibration data is collected in a relatively open environment with no blockages to the signal. Consequently, we can suppress the phase difference induced by bearings as
\begin{align*}
    W^\mathrm{sup}_t = W_t \odot \mathrm{conj}(\hat{W}_t)
\end{align*}
This leaves the remaining calibration phase $C$ in $W^\mathrm{sup}_t$. However, each WiFi measurement may have multiple reflected paths and hardware-centric Gaussian noise, which have not been adequately suppressed. However, we have two hints. One, reflections are inconsistent across different locations; two, averaging can suppress Gaussian noise. Hence, the best calibration estimate is the strongest remaining component in the suppressed $W^\mathrm{sup}_t$ measurements. We can leverage Principle Component Analysis to extract this strongest component in our calibration data as

\begin{align*}
    W^\mathrm{flat}_t &= \mathrm{flatten}(W^\mathrm{sup})t), \quad W^\mathrm{flat}_t \in \mathbb{C}^{4 N_f} \\
    \mathbb{W} &= \begin{bmatrix}  W^\mathrm{flat}_0 & W^\mathrm{flat}_1 & \cdots & W^\mathrm{flat}_T \end{bmatrix} \\ 
    \mathbb{U}, \mathbb{S}, \mathbb{V} &= \mathrm{SVD}(\mathbb{W}) \\
    \Phi^\mathrm{coarse} &= \angle \mathrm{reshape}(\mathbb{U}_0), \quad C^\mathrm{coarse}  = \exp(j\Phi^\mathrm{coarse})
\end{align*}
where `flatten' converts the matrix into a vector, SVD computes the full singular-value decomposition, `reshape' converts the vector back into a matrix of the original dimensions, and $\angle$ computes the phase of the complex numbers. $\mathbb{U}_0$ is the first and strongest principal component  of $\mathbb{W}$. However, as indicated, this calibration is a coarse estimate. The expectation is for the calibration matrix to consist of unit-norm \textit{elements}, but the principle component, $\mathbb{U}_0$, is a unit-norm \textit{vector}, violating this property. Hence, we need to re-project $C^\mathrm{coarse}$ onto a valid space of calibrations. To find a valid calibration,  $C^\mathrm{fine}$, that is close to $C^\mathrm{coarse}$, we note that $C^\mathrm{fine}$ must be orthogonal to the other vectors in $\mathbb{U}$,  so we try to find a fine-tuned calibration $\Phi^\mathrm{fine}$ which has the lowest norm when it is projected onto $\mathbb{U}_{[1:]}$.

\begin{align*}
    \Phi^\mathrm{fine} &= \min_\Phi ||\mathbb{U}_{[1:]}^T \,\mathrm{flatten}(\exp(j \Phi))||_2^2 \\
    C^\mathrm{fine} &= \exp(1j \Phi^\mathrm{fine})
\end{align*}
where $\mathbb{U}_{[1:]}$ is the orthogonal space, and we minimize for $\Phi$ by leveraging the Levenberg-Marquardt~\cite{kelley1999iterative} algorithm with an initialization of $\Phi^\mathrm{coarse}$. Via this fine-tuning process, we recover the wireless phase calibration matrix $C^\mathrm{fine}$. Later in Section~\ref{sec:eval}, we will evaluate the accuracy and versatility of this calibration across different WiFi frequency channels and hardware restarts.  

\subsection{Feature-extraction Node: Library of CSI processing tools} \label{sec:bearing}

The measurements from the `CSI Node' can be leveraged to measure signal path parameters like signal strength, angles of arrival or departure, the velocity of the transmitter, or locations of reflections in the environment. However, angular information can be readily measured from a single packet and provides a quick way to realize the advantages of WiFi measurements. In this vein, we open-source a large set of state-of-art bearing estimation tools~\cite{spotfi,  schmidtmusic} and CSI filtering techniques. 

\noindent \textbf{Usage:} Deploy \name's receiver on the robot and compensate the hardware as previously mentioned. We can set up \name's transmitter or monitor existing WiFi packets in the environment to measure the CSI measurements using the CSI-Node. Leveraging these CSI measurements, we can compute the angle of arrival (AoA) of a signal at the receiver (robot-side bearing) and the angle of departure of a signal at the transmitter (AP-side bearing) from a single WiFi signal received at the robot. These bearings would be in the robot's and the AP's local coordinate frames and are exposed via the \texttt{rf\_msgs/Bearing} ROS \textit{topic}. The `Feature-extraction' node also generates specialized visualizations to help debug the wireless channel, as discussed in Section~\ref{sec:usage}.
 
\noindent \textbf{Behind the scenes:} The `Feature-extraction' node is written in Python3, as this allows easy modification by users to support their specific sensing needs. It collects and sorts CSI messages and passes them to a data consumer object, collating the CSI measurements from a specific target device and using them to generate bearing estimates. We provide several example consumers which run a variety of AoA algorithms, from a very compute-efficient bearing-only algorithm that can be run in real-time to computationally heavy algorithms like SpotFi\cite{spotfi}.  Additionally, we note that CSI measurements can often fluctuate during motion because of dynamic reflections in the environment. In situations where we need to measure the direct signal path, these fluctuations can be detrimental. Hence, to obtain reflections-suppressed CSI measurements, we also implement an averaging technique from our prior work~\cite{arun2022p2slam}. Supporting math for these algorithms is provided with the documentation$^{\ref{fn:links}}$.

\begin{figure}
    \centering
    \includegraphics[width=0.9\linewidth]{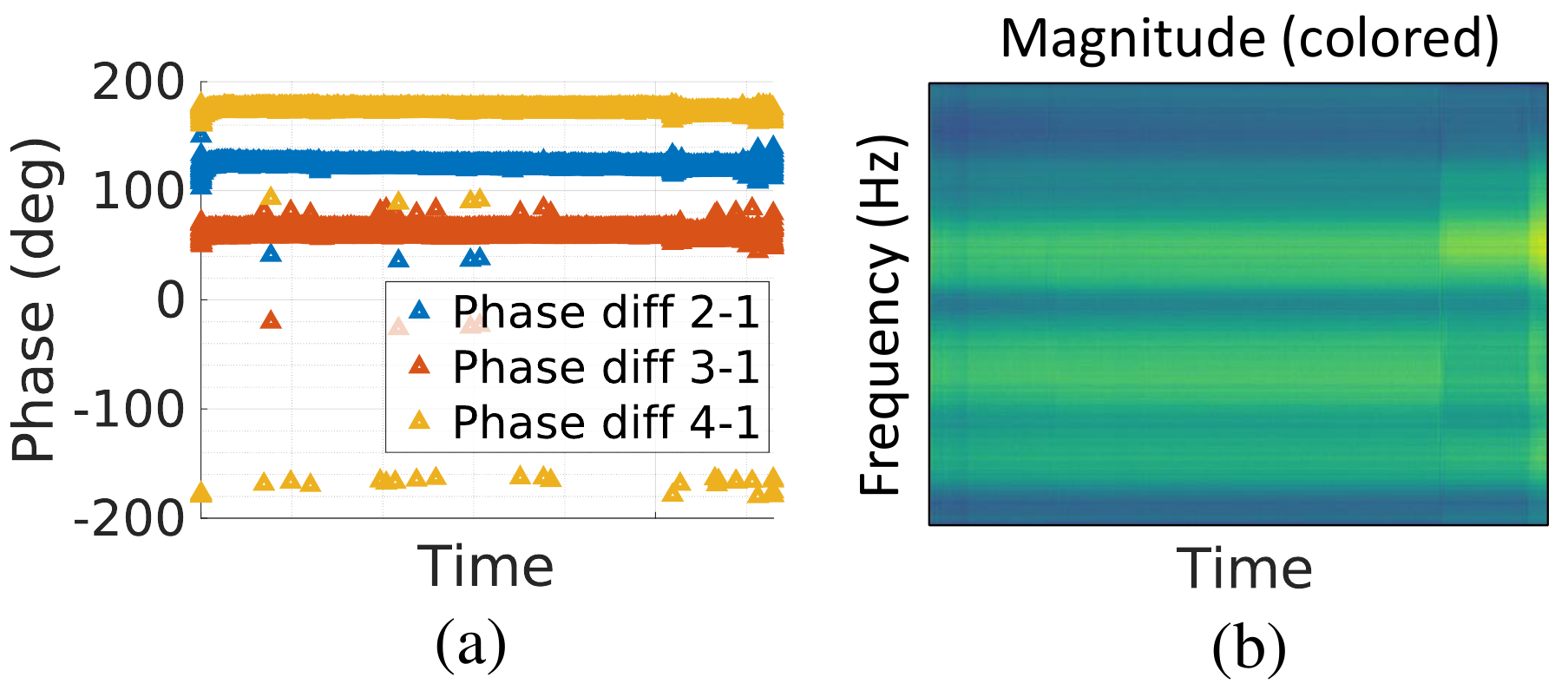}    
    \caption{\textbf{Stability of WiFi measurements:} WiFi measurements collected over 18 hours showcases \textbf{(a) phase stability}: CSI phase at DC frequency for 4 receiver antennas and, \textbf{(b) magnitude stability}: CSI spectrogram across all transmitted frequencies for an $80$ MHz bandwidth signal.} 
    \label{fig:eval1}
\end{figure}

\subsection{Extending \name to other CSI toolkits} \label{sec:other}

The CSI Node and Feature-extraction Node form the cornerstones of \name. The CSI Node interfaces \name with the underlying hardware, whereas the Feature-extraction Node open-source state-of-the-art post-processing algorithms for the WiFi CSI measurements. \name currently extends the Nexmon CSI extraction toolkit~\cite{ubilocate}, making specific API calls to configure the underlying hardware. However, we can follow a similar design template to adapt \name to other CSI toolkits~\cite{gringoli2022ax, jiang2021eliminating} as well. Additionally, we encourage future hardware platforms and CSI-extraction toolkits to prioritize exposing data via a UDP socket. Consequently, these newer platforms can enjoy ROS support with minimal changes to \name's `Sensor Config' and `Decode data' modules. 
\section{Verification Steps} \label{sec:eval}

\begin{figure*}
    \centering
    \includegraphics[width=0.9\textwidth]{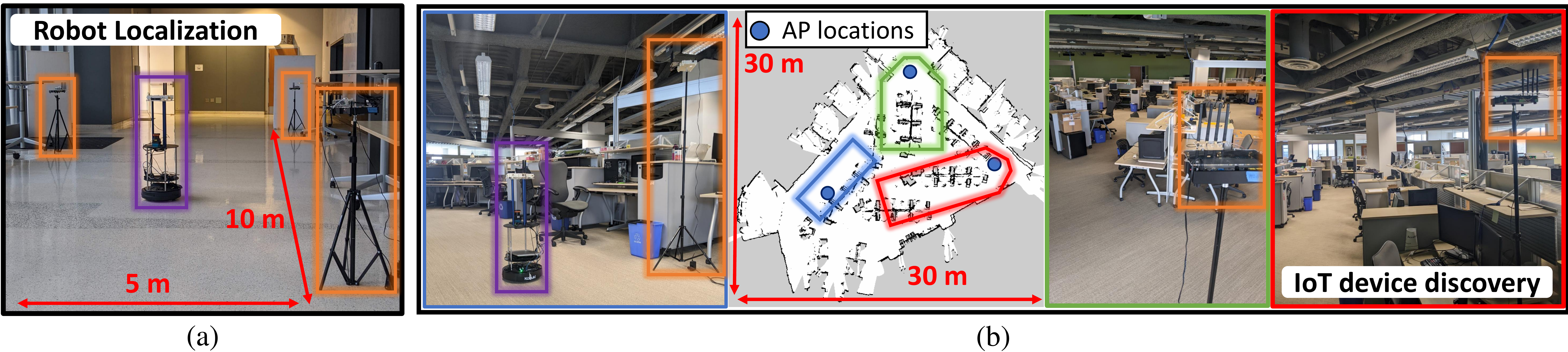}
    \caption{Testing Environments for case-studies  (a) Kidnapped robot problem and (b) IoT Localization. The purple and orange boxes show the robot (with \name's receiver) and \name's transmitter. In (b), the colored photo frames (blue, green and red) correspond to the camera viewpoint in the 2D top-down map.}
    \label{fig:envs}
\end{figure*}

In the following section, we will use the Asus AC86U~\cite{asus} (henceforth just WiFi sensor) as an example to verify the various aspects of \name. We deploy this WiFi sensor on a Turtlebot 2 platform~\cite{turtlebot2} equipped with a Hokuyo Lidar~\cite{hoku} and Realsense Camera~\cite{realsense}. We collect all data on a Thinkpad 13" Laptop~\cite{lenovo}. Specifically, we care about the stability of our signal measurements, the efficacy of our wireless calibration techniques, and the performance of the bearing estimation and automated channel switching feature. These sets of experiments provide a guideline for testing a WiFi sensor. If the specified Asus hardware is used, the reader should expect similar results as showcased. However, \name can be easily extended to other hardware systems as well, in which case, the following sets of verification can be performed for the new sensor.

\subsection{Measurement stability}

First, we ensure our sensor provides stable measurements over time, unaffected by the board's temperature differences and minor disturbances to the setup (as may be expected when the sensor is deployed on the robot). Two WiFi sensors were placed in an environment. One was set up to transmit data at 1 Hz and another to receive this transmitted data (see Section~\ref{sec:usage}), which were subsequently stored in a \textit{bag} file. 

In Figure~\ref{fig:eval1}(a), we showcase the phase stability across antennas at the center frequency over an $18$ hour run. We plot the same for the magnitude of the received signal (in color intensity) across the different frequencies (in the y-axis) for a single antenna over time (in the x-axis) in Figure~\ref{fig:eval1}(b). These plots show little variation in the phase and magnitude measurements over time, indicating that the CSI measurements are reliable and consistent.   

\subsection{Calibration efficacy}

Next, we must effectively correct hardware biases and offsets to measure accurate bearings. Specifically, we observe that without appropriate calibration, our median bearing errors can be $115^\circ$. However, post calibration, we can expect median bearing errors of $5.3^\circ$. Hence we provide a wireless calibration system as described in Section~\ref{sec:calib} as a core part of \name. However, this technique must work consistently for different WiFi channels and upon device reset. 

\begin{figure}
    \centering
    \includegraphics[width=0.9\linewidth]{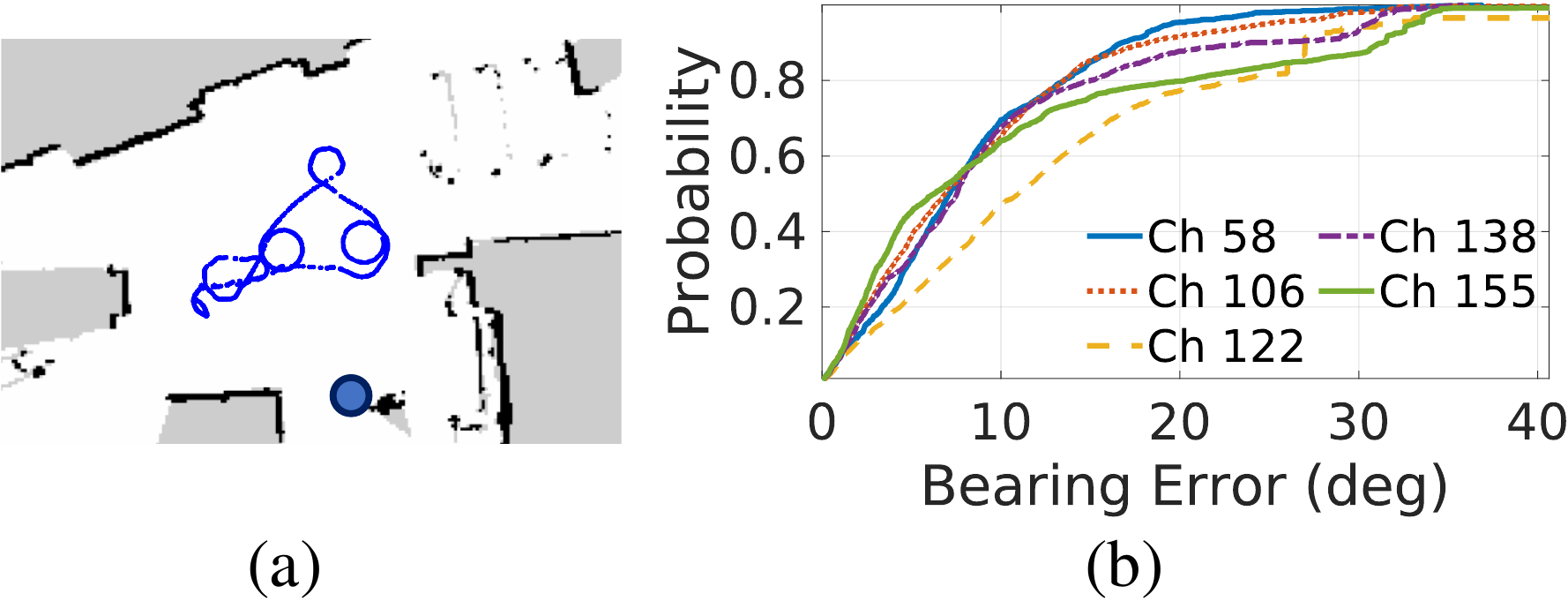}
    \caption{\textbf{Consistency and accuracy of calibration} \textbf{(a)} Robot path and transmitter location (blue circle) to calibrate the robot's WiFi sensor in $3 \times 3$ m, \textbf{(b)} Bearing errors when calibrated across different channels, after reset and channel switching}
    \label{fig:calib_fig}
\end{figure}

We collect the data as explained in Section~\ref{sec:calib} to confirm the calibration performance. We place an additional sensor in the environment sending beacon packets and another on the robot receiving them. We run the robot in a random pattern, collecting CSI data within \textit{bag} files, as shown in Figure~\ref{fig:calib_fig}(a). This is additionally done on different WiFi channels. Next, we re-run these experiments in different environments (shown in Figure~\ref{fig:envs}) to test the calibration efficacy. 

Figure~\ref{fig:calib_fig}(b) shows bearing errors across different channels after the APs were reset post collecting the required training data. For this experiment, we measure calibration for different channels and test the quality of these calibrations by collecting a test dataset after restarting the AP by characterizing the bearing accuracy. We observe similar performance across all the $80$ MhZ WiFi channels~\cite{ieee2007ieee}. Additionally, note that these calibration properties are specific to the Asus hardware used in this verification process. The wireless calibration procedure provided is agnostic to hardware, but it is imperative to take these verification steps when working with different hardware.     

\begin{figure*}
    \centering
    \includegraphics[width=\linewidth]{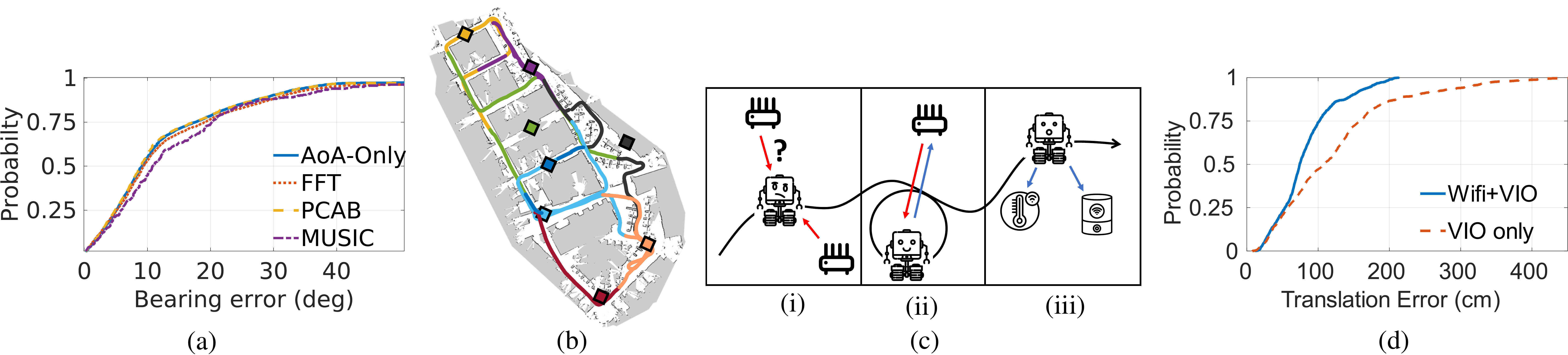}
    \caption{\textbf{(a)} Bearing estimation accuracy across algorithms, \textbf{(b)} \name can automatically detect and lock on to the nearest AP (labeled by color) while exploring the environment. \textbf{(c)} \textbf{Case-study overview}: \textbf{(i)} Lost robot can leverage \name to localize itself using the AP's in the environment. \textbf{(ii)} Drift during loop closures can be corrected using two-way bearings~\cite{arun2022p2slam}. (ii) IoT devices can be localized in the environment by leveraging \name. \textbf{(d)} Absolute Trajectory error when incorporating WiFi-bearings compared with Kimera~\cite{rosinol2021kimera}} 
    \label{fig:eval2}
\end{figure*}

\subsection{Bearing Estimation accuracy of open-sourced algorithms} 

Post calibration, we have corrected phase measurements which can be used for bearing estimation. In the previous section, we presented some bearing results to showcase calibration efficacy. However, we glossed over the specific algorithms used. As discussed in Section~\ref{sec:bearing}, we open source implementation of prior bearing estimation techniques~\cite{spotfi, schmidtmusic}. More details of these algorithms can be found in our documentation of the Feature-extraction node$^{\ref{fn:links}}$. The accuracy of these algorithms is shown in Figure~\ref{fig:eval2}(a). In these experiments, we ensure to include only received packets that have received signal strength stronger than $-65$ dB. This allows us to reject outliers. However, it is important to note that this threshold is a hardware-specific number, and other hardware systems may require different thresholds.  

\subsection{Feature: Automated channel switching}
We test the channel switching feature as explained in Section~\ref{sec:usage} in an enterprise network deployed in our university building in Figure~\ref{fig:eval2}(b). The robot traverses twice around the environment, shown in overlapping trajectory, in a single run. The colors represent the different APs deployed as part of the enterprise network, and the colors along the path indicate the specific AP \name's sensing channel is tuned into. The figure showcases that at different robot locations, \name appropriately tunes to the nearest AP, ensuring that the robot can use WiFi-sensing packets from line-of-sight access points. Furthermore, channel switching also depends on the path taken in the environment, which is apparent from the different behaviors across the two traversals. \name prioritizes continued interaction with the current AP rather than frequent AP switching to reduce downtime and consistency of data collection.

\section{Case-Studies} \label{sec:cases}

Next, we present three case studies to showcase the applicability of \name's various features. The corresponding Python3 scripts to get started with \name and these case studies are provided within the `Feature-extraction' node. 

\subsection{Localization to combat Kidnapped Robot Problem}

In the kidnapped robot problem, a robot carried to an arbitrary location in the environment is lost as it fails to re-localize within the global map. This is an additional challenge in GPS-denied scenarios where a robot does not have a global location estimate, which is common in indoor scenarios. However, some works~\cite{neto2022analysis} have used WiFi estimates to tackle this problem by providing global location estimates. 

\name can be leveraged to solve the kidnapped robot problem in indoor settings by leveraging existing deployed WiFi infrastructure as shown in Figure~\ref{fig:eval2}(c, i). By configuring \name's receiver to sense all WiFi packets in the environment, we can collect WiFi measurements from all nearby APs. This is made easier by the provided automated channel-switching feature. Next, \name can estimate the robot's bearing in the AP's frame of reference by leveraging the angle of departure (red arrows). These AP-sided bearings can be subsequently used to triangulate the robot. We need prior knowledge of the AP's location, antenna array geometry, and calibration metrics (which can be wirelessly computed using the `Feature-extraction' node) to deploy this application. As a test, we deploy 4 access points in a $5 \times 10 $ m environment (Figure~\ref{fig:envs}(a)) and observe a median localization accuracy of $1.2$ m.  

\subsection{Odometry-correction in SLAM systems}

Erroneous sensor measurements made by odometry sensors, cameras, or Lidars can create drift in a robot's predicted location. These drifts are corrected by applying ``\textit{loop closures}'', where a robot leverages the intuition that it must predict similar location estimates when revisiting previously seen spaces. However, due to perceptual aliasing~\cite{lajoie2019modeling}, different locations in the map may look similar, creating ambiguity when detecting loop closures. However, recent works~\cite{arun2022p2slam, hashemifar2019augmenting} have looked at fusing WiFi-bearing measurements to address the problems with visual loop closures. 

Again, \name can be leveraged to solve the issues with sensor drift during loop-closure as shown in Figure~\ref{fig:eval2}(c, ii). We can configure \name's receiver at the robot to collect WiFi measurements from APs in the wild. Subsequently, the measured robot-sided and AP-sided bearings from these WiFi measurements (red and blue arrows) can be fused with robot odometry in a factor graph to correct for drifts. Sensor fusion is made easy by \name as WiFi-based bearing measurements are readily available as ROS topics and time-synchronized with other sensor measurements on the robot. In Fig. \ref{fig:eval2}(d), we compare WiFi-based bearings fused with odometry from Visual-Inertial odometry (WiFi+VIO, no loop-closures) and a state-of-the-art Visual-inertial SLAM system, with loop closures enabled~\cite{rosinol2021kimera} (VIO only). We observe a $30\%$ reduction in median error due to improved robustness to visual aliasing.

\subsection{IoT device mapping to aid device management}

Finally, instead of localizing a robot in an environment (which is carrying a WiFi transceiver), we can consider the reverse problem of localizing WiFi transceivers in an environment. This is an important problem for IoT device management~\cite{chen2013localization}, where localizing various devices in a large space within a common map is necessary to aid building managers in maintaining the different IoT devices. Alternatively, recent works~\cite{sharma2022lumos} have looked at localizing, potentially rogue, IoT devices to ensure the privacy and security of users. 

\name can be easily leveraged to localize all WiFi IoT devices in a given space, as shown in Figure~\ref{fig:eval2}(d, iii). A robot deployed with \name's receiver, scanning all channels and enabling an all-pass MAC filter, can capture WiFi measurements from all WiFi-based IoT devices in space. Subsequently, the robot-sided bearing measurements (blue arrows) over multiple robot locations can be leveraged to triangulate the IoT devices. As a demonstration, we deploy \name's receiver on the robot and collect CSI from 3 WiFi transmitters deployed in a $10 \times 5$ m environment. Next, by utilizing the bearing estimates provided by \name, we can localize these devices with a location accuracy within $2$ m. The real-time operation of this case study is provided as a supplementary video$^{\ref{fn:video-link}}$. 

\section{Discussion and Future Work}

\name's primary deliverable is to make accurate WiFi-based sensing widely accessible for robotics use cases. Consequently, this work largely focuses on furnishing the calibrated physical Wifi channel measurements and bearings as a key feature to extract from these measurements. These WiFi-based bearings can be applicable for correcting errors accrued in online SLAM, re-localizing robots after sudden changes to their locations, or localizing WiFi transceivers in the environment for IoT management. However, the larger motivation of this work is to facilitate exploring the advantages of RF-based sensing within robot systems. This could be for robot exploration, collision avoidance, or motion planning. Additionally, we would like \name to serve as a template for future networking researchers building CSI-extraction tools to continue to provide support for newer WiFi protocols, with an immediate future step of extending support to upcoming 802.11ax WiFi protocols~\cite{gringoli2022ax, jiang2021eliminating}.

\newcommand{\ti}{\theta_i}

\bibliographystyle{IEEEtran}
\bibliography{references}

\end{document}